\pdfoutput=1
\documentclass[conference]{IEEEtran}
\usepackage{color}
\usepackage{times}
\usepackage{ifpdf}
\usepackage{url}
\ifpdf
  \usepackage[pdftex]{graphicx}
\else
  \usepackage[dvips]{graphicx}
\fi

\begin{document}

\title{Improving the Quality of Non-Holonomic Motion by Hybridizing C-PRM Paths}

% avoiding spaces at the end of the author lines is not a problem with
% conference papers because we don't use \thanks or \IEEEmembership

\author{\authorblockN{Itamar Berger\authorrefmark{1}\authorrefmark{3},
Bosmat Eldar\authorrefmark{1}\authorrefmark{3}, Gal
Zohar\authorrefmark{1}\authorrefmark{3}, Barak
Raveh\authorrefmark{1}\authorrefmark{2}\authorrefmark{4} and Dan
Halperin\authorrefmark{1}\authorrefmark{4}}
\authorblockA{\authorrefmark{1}School of Computer Science, Tel-Aviv University, Tel-Aviv, Israel}
\authorblockA{\authorrefmark{2}Department of Microbiology and Molecular Genetics, The Hebrew University, Jerusalem, Israel}
\authorblockA{\authorrefmark{3}I.B., B.E. and G.Z. contributed equally to this work}
\authorblockA{\authorrefmark{4}Email: barak,danha@post.tau.ac.il} }

\maketitle

%\begin{abstract}
%The abstract goes here.
%\end{abstract}

\IEEEpeerreviewmaketitle

\let\thefootnote\relax\footnotetext{
This abstract appeared in the workshop on Motion Planning: From
Theory to Practice, as part of the Robotics: Science and Systems
(RSS) conference, Zaragoza, 2010}

\section{Introduction}
Sampling-based motion planners are an effective means for generating
collision-free motion paths. However, the quality of these motion
paths, with respect to different quality measures such as path
length, clearance, smoothness or energy, is often notoriously low.
This problem is accentuated in the case of non-holonomic
sampling-based motion planning, in which the space of feasible
motion trajectories is restricted. In this study, we combine the
C-PRM algorithm by Song and Amato~\cite{Song01} with our recently
introduced path-hybridization approach~\cite{Raveh2010}, for
creating high quality non-holonomic motion paths, with combinations
of several different quality measures such as path length,
smoothness or clearance, as well as the number of reverse car
motions. Our implementation includes a variety of code optimizations
that result in nearly real-time performance, and which we believe
can be extended with further optimizations to a real-time tool for
the planning of high-quality car-like motion.

\subsection{C-PRM} 
\begin{figure}[!t]
  \centering
  \includegraphics[width=10cm]{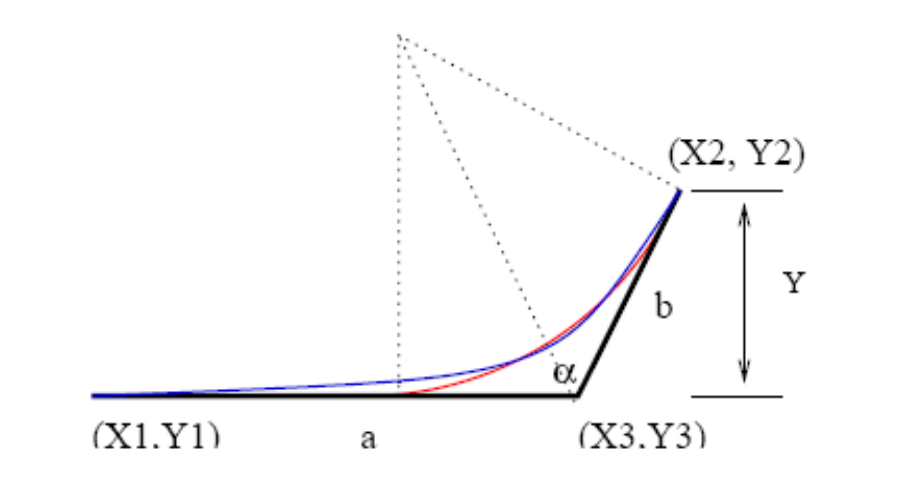}
  \caption{
  Moving from the \emph{control roadmap} (devoid of non-holonomic constraints) to the
  \emph{approximate roadmap} in the C-PRM algorithm for car-like motion
  planning. Figure is taken from from Fig. 2 in Song and Amato~\cite{Song01}.
  }
  \label{fig:1}       % Give a unique label
\end{figure}

The original C-PRM algorithm for planning car-like
motion~\cite{Song01} starts by building an initial \emph{control
roadmap}, a probabilistic roadmap devoid of any non-holonomic
constraints, in order to capture the coarse connectivity of the free
space. It then builds the \emph{approximate roadmap} by converting
groups of three nearby points to a valid car path made of straight
line segments and arcs that connect between midpoints of edges in
the control roadmap (Fig.~\ref{fig:1}). The C-PRM algorithm takes a
lazy approach in which collision checks and final refinements are
conducted at query time. Song and Amato showed that the C-PRM
algorithm is effective in generating car-like motion paths. Here, we
put a special emphasis on improving the quality of these paths.

\subsection{Path Hybridization for Improving Path Quality}
 We have recently introduced the path-hybridization approach~\cite{Raveh2010,Enosh08},
 in which an arbitrary number of input motion paths
are hybridized to an output path of superior quality, for a range of
path-quality criteria. The approach is based on the observation that
the quality of certain sub-paths within each solution may be higher
than the quality of the entire path. Specifically, we run an
arbitrary motion planner $k$ times (typically $k$=5-6), resulting in
$k$ intermediate solution paths to the motion planning query.  From
the union of all the edges and vertices in the intermediate paths we
create a single weighted graph, with edge weights set according to
the desired quality criterion. We then try to merge the intermediate
paths into a single high-quality path, by connecting nodes from
different paths with the local planner, and giving the appropriate
weights to the new edges. Dijkstra's algorithm is used to find the
highest-quality path in the resulting Hybridization-Graph (H-Graph).

\begin{figure}[!]
  \centering
  \includegraphics[width=8cm]{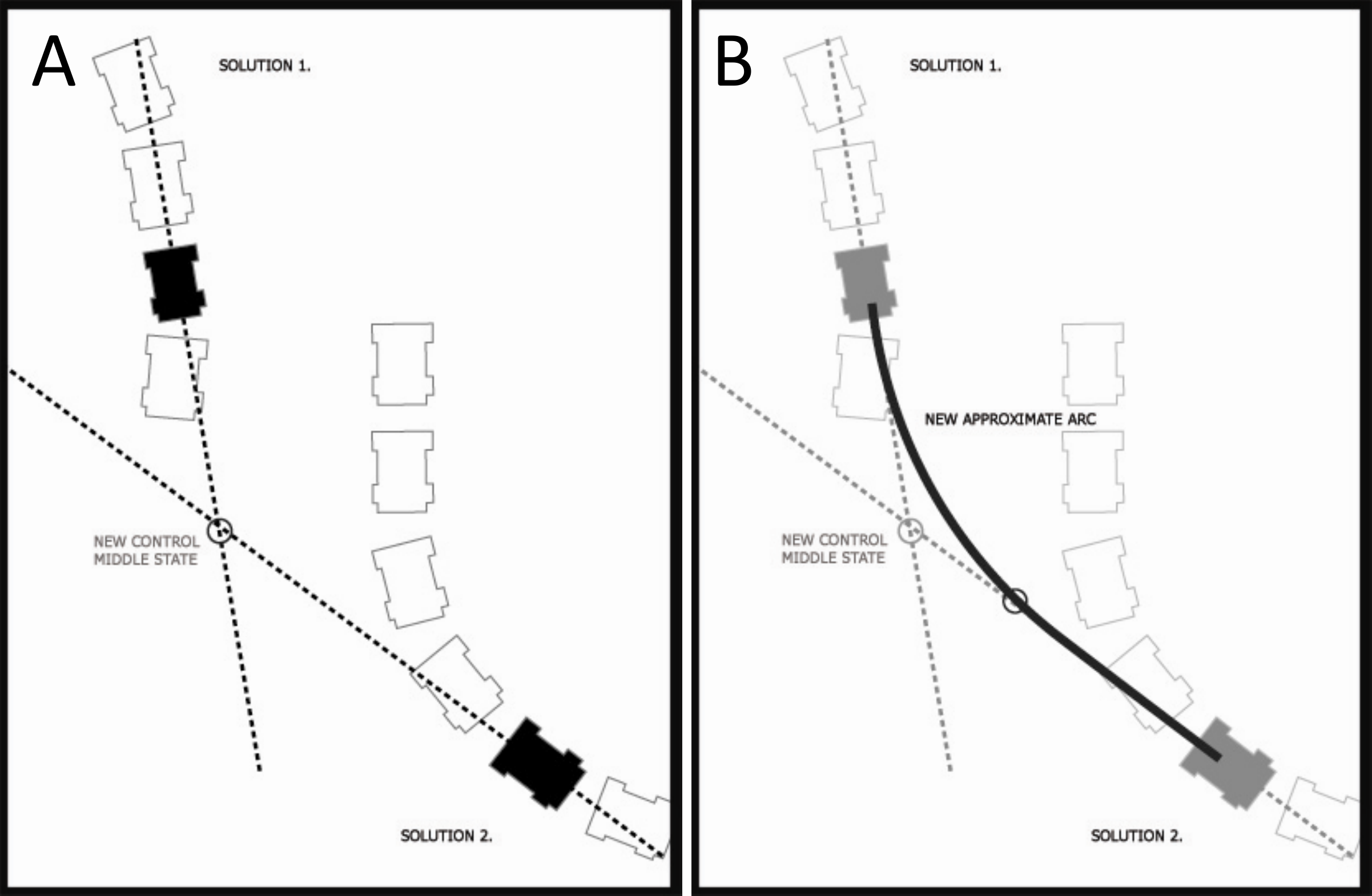}
  \caption{
  Connecting two car states by ``reverse engineering'' of the original C-PRM algorithm.
  \textbf{A} - Finding the intersection point between two car states.
  \textbf{B} - connecting the two states by an arc.
  }
  \label{fig:2}       % Give a unique label
\end{figure}

\section{C-PRM with path hybridization}
 While the path hybridization approach has been successfully tested over a range of
holonomic motion planning problems with many degrees of
freedom~\cite{Raveh2010,Enosh08}, its application to non-holonomic
motion planning is not trivial. In particular, whereas it is easy to
connect two nearby configurations in the case of holonomic motion
(for example, by linear interpolation between the two
configurations), it is not possible to simply interpolate between
two states of non-holonomic motion planning, due to the restriction
on the set of possible paths. However, we observed that we can
simply reverse engineer the original approach taken by C-PRM for
car-like motion planning, and go from the approximate roadmap (that
is made of arcs and line segments) to the control roadmap (the
coarse roadmap that does not include non-holonomic constraints),
instead of working the other way around (Fig.~\ref{fig:2}). This
allowed us to use the path-hybridization approach in a non-holonomic
setting, to generate car-like motion paths of high quality (for
example, see Fig. 3).

\subsection{Implementation} 
\begin{figure}[!t]
  \centering
  \includegraphics[width=9cm]{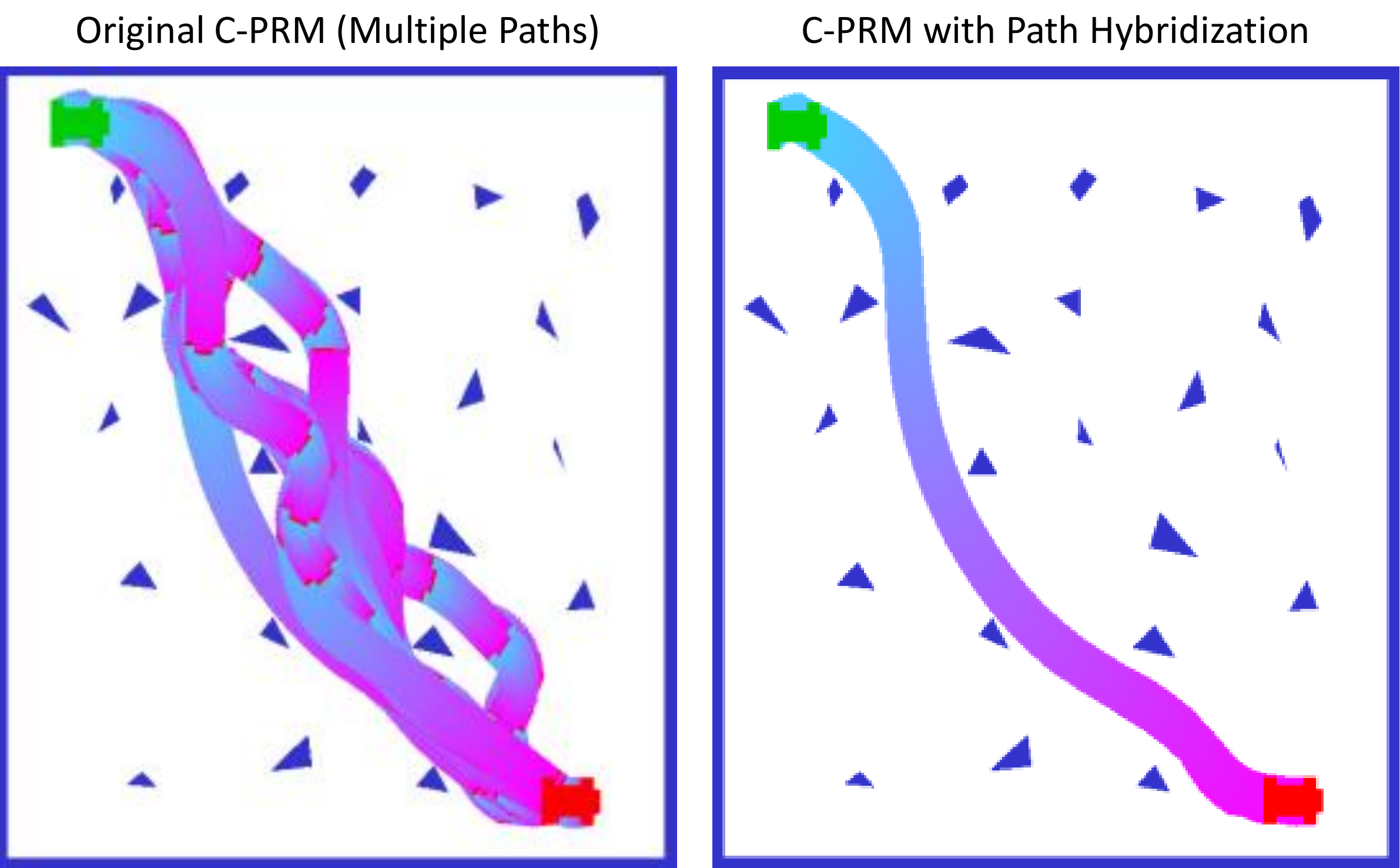}
  \caption{
  Experimental results of running C-PRM with Path Hybridization. The left panel shows an overlay
   of six different motion paths that were generated by the original C-PRM algorithm. The right
   panel shows the hybridization of these paths by the path hybridization algorithm to obtain a
   shorter path (as described in the text).
  }
  \label{fig:3}       % Give a unique label
\end{figure}

We have implemented the C-PRM algorithm
and C-PRM with path hybridization within the framework of the OOPSMP
motion planning package~\cite{PlaBekKav2007:ICRA_OOPSMP}. Our
implementation supports the combination of a wide range of path
quality criteria (length, smoothness, clearance, as well as the
number of reverse car motions). We also include a variety of code
optimizations, and our experiments indicate close to real-time
performance, which we believe could be even further improved by
further reasonable efforts. In Fig.~\ref{fig:3} we include an
illustrative example of motion planning for a car-like vehicle using
our implementation of both the original C-PRM algorithm for car-like
motion planning, and the C-PRM algorithm combined with path
hybridization. As noted in Raveh \emph{et al.}~\cite{Raveh2010},
path hybridization incurs an additional running time cost due to the
need for multiple runs of the C-PRM algorithm, but yields motion
paths of superior quality, with large flexibility in terms of the
quality criteria applied. Further details and a movie are included
in
\url{http://acg.cs.tau.ac.il/courses/workshop/spring-2009/final-projects/non-holonomic-motion-planner-project}.

\section{Conclusions}
The problem of finding high-quality motion paths is of prime
importance. Due to the constrained nature of non-holonomic motion,
it is more difficult to plan high-quality non-holonomic motion paths
than holonomic ones. We have designed and implemented the first
application of the path hybridization approach to non-holonomic
motion planning in the rather simple setting of car-like motion,
showing promising results. In order to hybridize non-holonomic
paths, we had to develop a non-holonomic local planner to connect
nearby states. Developing similar local planners for other
non-holonomic vehicles would allow the extension of the path
hybridization approach for a wide range of non-holonomic problems.

\bibliographystyle{IEEEtran}
\bibliography{IEEEabrv,RSS_references}

\end{document}